\documentclass[10pt,twocolumn,twoside]{IEEEtran}
\IEEEoverridecommandlockouts
\usepackage{epsfig}
\usepackage{graphicx}
\usepackage{amsmath}
\usepackage{amssymb}
\usepackage{multirow}
\usepackage{subfigure}
\usepackage{algorithm}
\usepackage{algorithmic}
\usepackage{subfigure}
\usepackage{color}

\DeclareMathOperator*{\argmin}{arg\,min}
\usepackage[breaklinks=true,bookmarks=false]{hyperref}

\begin{document}
%




\title{Text-Attentional Convolutional Neural Network \\ for Scene Text Detection}

\author{Tong~He,
~Weilin~Huang,~\IEEEmembership{Member,~IEEE,}
~Yu~Qiao,~\IEEEmembership{Senior~Member,~IEEE,}
        and~Jian~Yao,~\IEEEmembership{Senior~Member,~IEEE}

\thanks{Tong He, Weilin Huang and Yu Qiao are with Shenzhen Institutes of Advanced Technology, Chinese Academy of Sciences, Shenzhen, China.\protect\\
  Tong He and Jian Yao are with School of Remote Sensing and Information Engineering, The Wuhan University, Wuhan, China.\protect\\
  Weilin Huang and Yu Qiao are also with Multimedia Laboratory, the Chinese University of Hongkong, Shatin, N.T., Hongkong.\protect\\
Corresponding Author: Weilin Huang (e-mail:wl.huang@siat.ac.cn)}

}



\maketitle

\begin{abstract}
%

 Recent deep learning models have demonstrated strong capabilities for classifying text and non-text components in natural images. They extract a high-level feature computed globally from a whole image component (patch), where the cluttered background information may dominate true text features in the deep representation. This leads to less discriminative power and poorer robustness. In this work, we present a new system for scene text detection by proposing a novel Text-Attentional Convolutional Neural Network (Text-CNN) that particularly focuses on extracting text-related regions and features from the image components. We develop a new learning mechanism to train the Text-CNN with multi-level and rich supervised information, including  text region mask, character label, and binary text/non-text information. The rich supervision information enables the  Text-CNN  with a strong capability for discriminating ambiguous texts, and also increases its robustness against complicated background components. The training process is formulated as a multi-task learning problem, where low-level supervised information greatly facilitates main task of text/non-text classification.
In addition, a powerful low-level detector called Contrast-Enhancement Maximally Stable Extremal Regions (CE-MSERs) is developed, which extends the widely-used MSERs by enhancing intensity contrast between text patterns and background. This allows it to detect highly challenging text patterns, resulting in a higher recall. Our approach achieved promising results on the ICDAR 2013 dataset, with a F-measure of 0.82, improving the state-of-the-art results substantially.

\end{abstract}


\IEEEpeerreviewmaketitle

\section{Introduction}
\label{Sec:Introduction}

 \begin{figure}
\centering
\subfigure[The confident scores by CNN \cite{Huang2014,Wang2012} (Yellow) and Text-CNN (Red)]{\includegraphics[width=8.5cm]{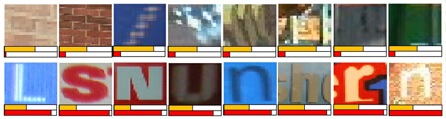}}
\subfigure[Image]{\includegraphics[height=6.5cm,width=2.8cm]{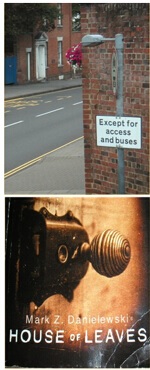}}
\subfigure[CNN \cite{Huang2014,Wang2012}]{\includegraphics[height=6.5cm,width=2.8cm]{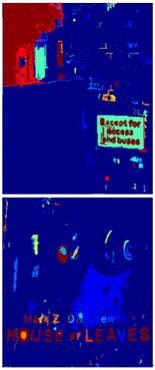}}
\subfigure[Text-CNN]{\includegraphics[height=6.5cm,width=2.8cm]{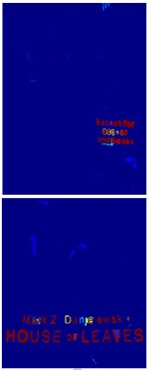}}

   \caption{Comparisons of confident maps/scores between the proposed Text-CNN and the CNN filter of~\cite{Huang2014,Wang2012}. (a) The confident scores of a number of text and non-text components: the Text-CNN in RED bars and CNN of~\cite{Huang2014,Wang2012} in YELLOW bars. (b-d): the Text-CNN in (d) shows stronger robustness against background components, and higher capability for discriminating ambiguous text components than the CNN of~\cite{Huang2014,Wang2012} in (c).}
\label{fig:main}
\end{figure}

 \IEEEPARstart {T}{ext} detection and recognition in natural images have received increasing attention in computer vision and image understanding, due to its numerous potential applications in image retrieval, scene understanding, visual assistance, etc.  Though tremendous efforts have recently been devoted to improving its performance, reading texts in unconstrained environments is still extremely challenging and remains an open problem, as substantiated by recent literature~\cite{Jaderberg2015,Zhang2015,Huang2014,Yin2014}, where the leading performance on the detection sub task is of $80\%$ F-measure on the ICDAR 2011~\cite{Zhang2015}, and current result of unconstrained end-to-end recognition is only $57\%$ accuracy on the challenging SVT dataset~\cite{Jaderberg2015}. In this work, we focus on the detection task that aims to correctly localize exact positions of text-lines or words in an image. Text information in natural images may have significant diversity of text patterns in highly complicated background. For example, text can be in a very small size, low quality, or low contrast, and even regular ones can be distorted considerably by numerous real-world affects, such as perspective transform, strong lighting, or blurring. These pose fundamental challenges of this task where detecting correct character components is difficult.

%

 The Extremal Regions (ERs) or Maximally Stable Extremal Regions (MSERs) based  methods achieved the state-of-the-art performance on scene text detection \cite{Huang2014,Yin2014}. The ERs detector exhibits great advantage in detecting challenging text patterns, and often results in a good recall. However, low-level nature of the ERs or MSERs makes them easily to be distorted by numerous real-world affects and complicated background. This results in a huge number of non-text components, which may be many orders of magnitude larger than the number of true text components, e.g. over $~10^6$ ERs and $~10^3$ MSERs per image, compared to dozens of characters in an image \cite{Neumann2012}. Therefore, correctly filtering out these non-text components is critical to the success of such methods.

%



Many previous methods focus on developing hand-crafted  features based on a number of heuristic image properties (e.g. intensity variance, sharp information or spatial location) to discriminate text and non-text components~\cite{Yao2012,Neumann2012}. These low-level features inherently limit their generality on highly challenging text components. They also reduce the robustness against text-like components, which often have similar low-level properties as the true texts, such as bricks, windows or leaves. These facts pose main challenge of current scene text detection systems, and severely harm their performance in both precision and recall.

Recently, a number of deep models have been developed for text component filtering/classification \cite{Huang2014,Jaderberg2014,Wang2012}, or word/character recognition~\cite{He2015,Jaderberg2014,Bissacco2013,Wang2012},  by leveraging advances of deep image representation. Huang \emph{et  al.}~\cite{Huang2014} applied a two-layer CNN with a MSERs detector for text detection. Jaderberg \emph{et al.}~\cite{Jaderberg2014} and Wang \emph{et al.}~\cite{Wang2012} employed CNN models for both detection and recognition tasks. These text classifiers, building on conventional CNN, are generally trained with binary text/non-text label, which is relatively less informative to learn a meaningful text feature. Furthermore, they  compute a high-level deep feature globally from an image patch for discriminating text and non-text. Such feature commonly includes a large amount of background information that possibly dominates in the representation, so as to reduce its discriminative power and robustness. As shown in Fig.~\ref{fig:main} (a), a number of background components are classified incorrectly as text by the CNN used in~\cite{Huang2014,Wang2012}, while some ambiguous text components may be misidentified as the non-text ones.

Our goal is to design a text CNN model that particularly focalizes to text-related regions and specific characteristics of text within an image component. In particular, this is achieved by training a CNN with more informative supervised information, such as text region mask, character label and binary text/non-text information. We show that such meaningful supervision would be greatly helpful to make a reliable binary decision on text or non-text. To realize this capability and incorporate additional supervised knowledge, we propose a novel Text-Attentional Convolutional Neural Network (Text-CNN) for text component filtering. The Text-CNN is incorporated with a newly-developed Contrast-Enhanced MSERs (CE-MSERs) to form our text detection system. Our contributions are summarized as follow.


Firstly, we propose the Text-CNN that particularly focuses on extracting deep text feature from an image component. Applying a deep CNN for text/non-text classification is not new, but to our knowledge, this is the first attempt to design a CNN model specially for text-related region and text feature computing, by leveraging multi-level and rich supervised information. The additional low-level supervised information enables our Text-CNN with better discriminative power to identify ambiguous components, and stronger robustness against background, as shown in Fig.~\ref{fig:main}.


Secondly, we introduce a deep multi-task learning mechanism to learn the Text-CNN efficiently, where each level of the supervised information is formulated as a learning task. This allows our model to share learned features between multiple tasks, making it possible to learn more discriminative text features by using the additional low-level supervision. This greatly facilitates convergence of the main task for text/non-text classification.

Thirdly, we develop a new CE-MSERs detector, which extends
the widely-used MSERs method by enlarging the local contrast between text and background regions.  The CE-MSERs detector is capable of detecting highly ambiguous components which are often missed or confused by the original MSERs. We incorporate the CS-MSERs with our Text-CNN filter to form a new text detection system which achieves the state-of-the-art results on the ICDAR 2011 and ICDAR 2013 benchmarks.

The rest of paper is organized as follows. A brief review on related studies is given in Section
II. Then the proposed text detection system, including the Text-CNN model and CE-MSERs detector, is described in Section III. Experimental results are compared and discussed in Section IV, followed by conclusions in Section V.




\section{Related Work}

Existing work for scene text detection can be roughly categorized into two groups,
 sliding-window and connected component based methods~\cite{Ye2015}. The sliding-window methods detect text information by moving a multi-scale sub-window through all possible locations in an image~\cite{Jaderberg2014,Zhang2015,Wang2012,Kim2003,Chen2004,Hanif2009,Wang2011,Mishra2012,Pan2011}. Then a pre-trained classifier is applied to identify whether text information is contained within the sub-window. For example, Wang, \emph{et al.} \cite{Wang2011} explored Random Ferns classifier \cite{Ozuysal2007} with a histogram of oriented gradients (HOG) feature \cite{Dalal2005} for text detection. Similarly, Pan \emph{et al.} \cite{Pan2011} generated text confident map by using a sliding-window with a HoG feature and a WaldBoost \cite{Sochman2005} which is a boosted cascade classifier. The main difficulties for this group of methods lie in designing a discriminative feature to train a powerful classifier, and managing computational flexibility by reducing the number of the scanning windows.



The connected component methods have achieved great success in text detection and localization~\cite{Yin2014,Huang2014,Yin2015,Huang2013,Kang2014,Li2014,Neumann2012,Yao2012,Epshtein2010,Neumann2011,Neumann2013,Yi2011,Chen2012,Sun2015}. They separate text and non-text information at pixel-level by running a fast low-level detector. The retained pixels with similar properties are then grouped together to construct possible text components. The ERs/MSERs~\cite{Matas2004,Nister2008} and Stroke Width Transform (SWT)~\cite{Epshtein2010} are two representative methods in this group. Extending from the original SWT, Huang \emph{et. al.}~\cite{Huang2013} proposed a Stroke Feature Transform (SFT), by incorporating important color cues of text patterns for pixel tracking. In~\cite{Huang2014,Yin2014,Kang2014,Neumann2011}, the MSERs detector has demonstrated strong capability for detecting challenging text patterns, yielding a good recall in component detection. In \cite{Li2014}, Characterness was proposed by incorporating three novel cues: stroke width, perceptual divergence, and  HoG at edges. Recently, Jaderberg \emph{et al.}~\cite{Jaderberg2015} applied the EdgeBox proposals ~\cite{Zitnick2014} for generating text components. Sun, \emph{et al.}  proposed a robust text detection system by developing a color-enhanced contrasting extremal region (CER) for character component detection, followed by a neural networks classifier for discriminating text and non-text components \cite{Sun2015}.

In order to handle multi-oriented text lines, a two-level approach was proposed for sequentially detecting component candidates and text chain candidates \cite{Yao2012}. Then two filters building on Random Forest \cite{Breiman01} were developed to effectively classify corresponding two-level candidates. Extending from this work, Yao \emph{et al.} \cite{Yao2014} proposed a unified framework for multi-oriented text detection and recognition, where same features were used for both tasks. Then a new dictionary search approach was developed to improve recognition performance. In \cite{Yin2015}. Yin \emph{et al.} presented a new multi-orientation text detection system by proposing a novel text line construction method. They designed  a number of sequential coarse-to-fine steps to group character candidates based on a pre-trained adaptive clustering.

The connected component based methods exhibit great advantage in speed by fast tracking text pixels in one pass computation, with complexity of $O(N)$. However, low-level nature of these methods largely limits their capability, making them poorer robust and discriminative.  Therefore, a sophisticated post-processing method is often required to deal with large amount of generated components, which causes main challenge of this group of methods.

A powerful text/non-text classifier or component filter is critical
to success of both the sliding-window and connected component based methods. Huge efforts have been devoted to developing an efficient hand-crafted feature that could correctly capture discriminative characteristics of text. Chen and Yuille~\cite{Chen2004} proposed a sliding-window method by using the Adaboost classifiers trained on a number of low-level statistical features. Similarly, Zhang \emph{et. al.}~\cite{Zhang2015} proposed a symmetry-based text line detector that computes both symmetry and appearance features based on heuristic image properties. For the connected component approaches, in~\cite{Epshtein2010}, a number of heuristic rules were designed to filter out the non-text components generated by the SWT detector. Extending from this framework, a learning based approach building on Random Forest~\cite{Bosch2007} was developed by using manually-designed histogram features computed from various low-level image properties~\cite{Yao2012}. To eliminate the heuristic procedures, Huang \emph{et. al.}~\cite{Huang2013} proposed two powerful text/non-text classifiers, named Text Covariance Descriptors (TCDs), which compute both heuristic properties and statistical characteristics of text stokes by using covariance descriptor \cite{Tuzel2006}. However, low-level nature of these manually-designed features largely limit their performance, making them hard to discriminate challenging components accurately, such as the ambiguous text patterns and complicated background outliers.

Deep CNN models are powerful for image representation by computing meaningful high-level deep features~\cite{Krizhevsky2012,Girshick2014,Simonyan2015,Szegedy2015}. Traditional CNN network (such as the well-known LeNet) has been successfully applied to text/document community for digit and hand-written character recognition~\cite{LeCun1989,LeCun1998}. Recently, the advances of deep CNN have also been adopted to design the challenging scene text detection/recognition systems \cite{Huang2014,Jaderberg2014,Wang2012,Jaderberg2015}. A CNN model was employed to filter out non-text components, which are generated by a MSERs detector in~\cite{Huang2014} or the Edgebox in~\cite{Jaderberg2015}, while Wang \emph{et al.}~\cite{Wang2012} and Jaderberg \emph{et. al.}~\cite{Jaderberg2014} applied a deep CNN model in the sliding-window fashion for text detection. Gupta \emph{et al.} developed a new Fully-Convolutional Regression Network (FCRN) that jointly performs text detection and bounding-box regression \cite{Gupta2016}. Though these deep models have greatly advanced previous manually-designed features, they mostly compute general image features globally from a whole image component/patch mixing with cluttered background, where background information may be computed dominantly in feature learning processing. They only use relatively simple text/non-text information for training, which is significantly insufficient to learning a discriminative representation.

Our work is closely related to recently proposed
approach by Huang \emph{et. al.}~\cite{Huang2014}, who incorporated a two-layer CNN with a MSERs detector for text detection. Our method improves upon this approach by developing a novel Text-CNN classifier and an improved CE-MSERs detector, which together lead to a significant performance boost. Our Text-CNN model is specially designed to compute discriminative text features from general image components, by leveraging more informative supervised information that facilitates text feature computing. This enables it with strong robustness against cluttered background information, and remarkably sets us apart from previous CNN based methods for this task.


\section{Proposed Approach}


The proposed  system mainly includes two parts: a Text-Attentional Convolutional Neural Network (Text-CNN) for text component filtering/classfication, and  a Contrast-Enhanced MSERs (CE-MSERs) detector for generating component candidates.

\subsection{Text-Attentional Convolutional Neural Network}
\begin{figure*}
\begin{center}
\includegraphics[height=5cm, width=16cm]{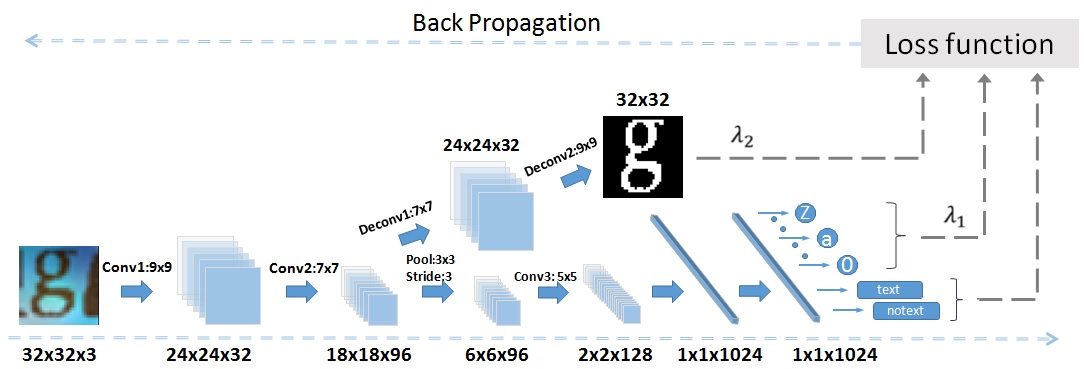}
\end{center}
   \caption{Structure of the Text-Attentional Convolutional Neural Network.}
\label{fig:text-cnn}
\end{figure*}

Given an ambiguous  image component, people can easily discriminate it in text or non-text, with much more informative knowledge about it, such as pixel-level text region segmentation and character information (e.g., `a', `b', `c', etc.). Such low-level prior information is crucial for people to make a reliable decision.
The text region segmentation allows people to accurately extract true text information from cluttered background, while the character label (assuming that people understand this language) helps them make a more confident decision on ambiguous cases.

Current text component filters are mostly learned with just binary text/non-text labels, which are insufficient to train a powerful classifier, making it neither robust nor discriminative. We wish to train a more powerful deep network for text task. We aim to `teach' the model with more intuitive knowledge about the text or character. These important knowledge include lower-level properties of the text, such as explicit locations of text pixels and labels of characters. Toward this, we propose the Text-CNN by training it with multi-level highly-supervised text information, including text region segmentation, character label and binary text/non-text information. These additional supervised information would `tell' our model with more specific features of the text, from low-level region segmentation to high-level binary classification. This allows our model to sequentially  understand $where$, $what$ and $whether$ is the character, which is of great importance for making a reliable decision.

However, training a unified deep model that incorporates multi-level supervised information is non-trivial, because different level information have various learning difficulties and convergence rates. To tackle this problem, we formulate our training process as a multi-task learning (MTL) \cite{Evgeniou2007} problem that treats the training of each task as an independent process by using one of the supervised information. Deep CNN model is well suited for the MTL by sharing features \cite{Bengio2013}. We formulate our model as follows.



\subsubsection{Problem Formulation}
Given total $N$ training examples, denoted as $\{(\textbf{x}_i, y_i)\}_{i=1}^N$, the goal of traditional CNN is trying to minimize
\begin{equation} \label{Eq:TraditionalCNN}
   \mathop{\argmin}_{\textbf{W}}{\sum_{i=1}^{N}{\mathcal{L}(y_i, f(\textbf{x}_i;\textbf{W})) + \Psi(\textbf{W})}}
\end{equation}
where $f(\textbf{x}_i;\textbf{W})$ is a function parameterized by $\textbf{W}$. $\mathcal{L}(\cdot)$ denotes a loss function, which typically chooses the hinge loss for classification task, and the least square loss for regression task.
The $\Psi(\textbf{W})$ is regularization term. The training procedure just tries to find a mapping function that connects the input image patch and (single-task) output labels (e.g., 0/1 for binary classification) without any extra information. Many existing CNN-based models follow this formulation by simply using the binary text/non-text information~\cite{Jaderberg2014,Huang2014,Wang2012}.

We observe that Eq.~(\ref{Eq:TraditionalCNN}) can be readily extended to the traditional MTL problem by maximizing overall loss of all tasks. However, the traditional MTL considers equal contribution or importance for all tasks. In our model, three tasks are significantly different. The lower-level text region learning and character label classification are much more complicated than the main task of text/non-text classification, leading to different learning difficulties and convergence rates. To this end, our goal is to optimize the main task with assistance of two auxiliary tasks. Therefore, directly applying the traditional MTL to our problem is non-trivial.

It has been shown that deep neural networks can benefit from learning with  related tasks simultaneously \cite{Bengio2013}. Recently, Zhang~\emph{et al.}~\cite{Zhang2014} developed an efficient task-constrained deep model for facial landmark detection, where additional facial attributes information are utilized to facilitate learning of the main detection task. Motivated from this work, we formulate our problem as a MTL problem with one main task ($\mathcal{L}^m$) and two auxiliary tasks ($\mathcal{L}^a$):
\begin{equation}\label{MTL_general}
\mathop{\argmin}_{\textbf{W}^m,\textbf{W}^a}\!\!{\sum_{i=1}^{N}\!{\mathcal{L}^m(y^{m}_{i}\!,\!f(\textbf{x}_i\!;\!\textbf{W}^{m})) } \!\!+\!\! \sum_{i=1}^{N}\!\!{\sum_{a\in A}\!\!{\lambda^a\mathcal{L}^a(y^a\!, \!f(\textbf{x}_i;\!\textbf{W}^a ))} }},
\end{equation}
where $\lambda^a$ and $\textbf{W}^a$ denote importance factor and model parameters of the auxiliary tasks, respectively. Regularization terms are omitted for simplification. In our model, the main text/non-text task and the auxiliary character label task are the classification problem, while learning of text region is a regression problem. Our three tasks can be further detailed as follows:
\begin{eqnarray}\label{MTL_details}
\mathcal{L}^{B}(y^b_i,f(\textbf{x}_i;\textbf{W}^b) &=& y_i^b \log(p(y_i^b|\textbf{x}_i;\textbf{W}^b), \\
\mathcal{L}^{L}(y^l_i,f(\textbf{x}_i;\textbf{W}^l)&=&  y_i^l \log(p(y_i^l|\textbf{x}_i;\textbf{W}^l),\\
\mathcal{L}^{R}(\textbf{y}^r_i,f(\textbf{x}_i;\textbf{W}^r)&=& \|\textbf{y}^r_i - f(\textbf{x}_i;\textbf{W}^r) \| ^2.
\end{eqnarray}

\begin{figure*}
\centering
\subfigure[Conventional CNN]  {\includegraphics[height=4cm,width=5cm]{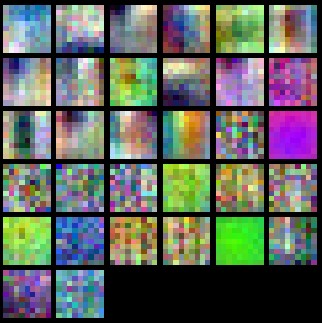}} \hspace{1em}
\subfigure[Text-CNN with only label task]  {\includegraphics[height=4cm,width=5cm]{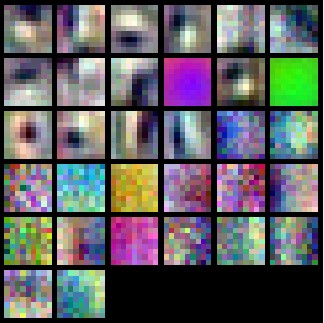}}\hspace{1em}
\subfigure[Text-CNN]  {\includegraphics[height=4cm,width=5cm]{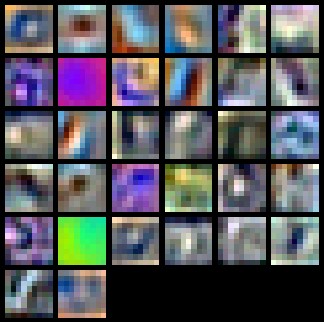}}
   \caption{Learned filters ($1$st-conv. layer) of  (a) the conventional CNN, (b) the Text-CNN with only label task, and (c) with both auxiliary tasks.}
\label{fig:filters}
\end{figure*}

Three tasks use the same input image patch, $\textbf{x}_i\in\mathbb{R}^{32 \times 32 \times 3}$.  The main differences between them are in output labels. $y^b_i=\{0,1\}\in\mathbb{R}^2$ (i.e. text/non-text), and $y^l_i=\{0...9,A...Z,a...z\}\in\mathbb{R}^{62}$ are the labels of main task and character label task, respectively. $\textbf{y}^r_i=\{0,1\}\in\mathbb{R}^{32 \times 32}$ is a binary mask with the same size of the input image (in 2D),  indicating explicit pixel-level segmentation of text. For the text region task, the model estimates probability of the text at each pixel: $f(\textbf{x}_i;\textbf{W}^r)=[0,1]\in\mathbb{R}^{32 \times 32}$. It minimizes the L2 distance between the ground truth mask and the estimated probability map. Notice that both auxiliary tasks are only explored in the training process. In the test process, only the main task works for text/non-text classification.

The details of Text-CNN are presented in Fig.~\ref{fig:text-cnn}. It includes three convolutional layers (with kernel size of $9\times9$, $7\times7$ and $5\times5$, respectively), followed by two fully-connected layers of 1024D. The second convolutional layer is followed by a max pooling layer with the pooling kernel of $3\times3$. The last fully-connected layer is followed the outputs of main text/non-text task (2D) and the character label task (62D). The text region regression is trained by using an additional sub network  branched from the second convolutional layer of the main network, before the non-invertible max pooling operation. It includes two deconvolutional layers \cite{Zeiler2014} duplicated from the 1st and 2nd convolutional layers of the main network, in an invert order.
The output of the (2nd) deconvolutional layer is a $32\times32$ probability map, corresponding to pixel locations of the input image.  $\lambda_1$ and $\lambda_2$ are two manually-setting parameters used to balance the two auxiliary tasks and main task.

\begin{figure}
\centering
\subfigure[Text components] {\includegraphics[height=4cm,width=4.2cm]{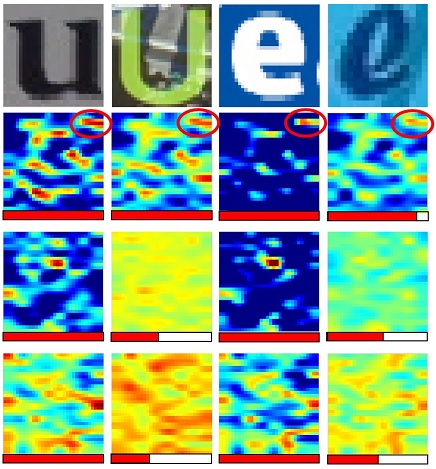}}
\subfigure[Non-text components] {\includegraphics[height=4cm,width=4.2cm]{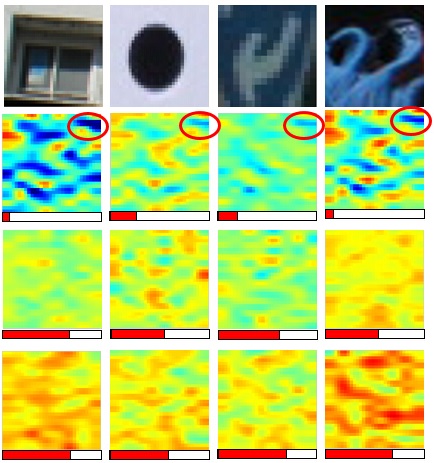}}
   \caption{Performance of the Text-CNN (2nd row), Text-CNN just with character label (3rd row), and the conventional CNN (4th row) on a number of ambiguous text and background components (1st row). The feature map ($32 \times 32$) is reshaped from the last fully-connected layer of each model, with its confident score under each map (in red bar, the higher score denotes higher probability to be the text). One of the key locations for discriminating text and non-text has been indicated by a red circle in each map.}
\label{fig:text-cnn-comp}
\end{figure}

The architecture of our model was designed carefully. First, the pooling layer was designed to balance computational complexity and model accuracy. In general, the pooling layer is developed to reduce model parameters and complexity, but at the cost of spatial information loss. Since we use small image patch (32 x32) as our model input (compared to the 227x227 input image in regular image classification tasks), our architecture does not need more pooling layers to reduce its spatial resolution. Second, since the pooling operation is non-invertible, so that it is difficult to apply it before the deconvolution layers starting at the 2nd convolution layer. Therefore, we do not use the pooling layer in the 1st convolution layer. In our experiments, removing the pooling operation in the 2nd convolution layer did not further improve the performance. But adding an additional pooling layer in the 3rd convolution reduced the accuracy, since the output map was already in a small size, 2$\times$2.

\subsubsection{Training the Text-CNN}

Our Text-CNN is trained by using stochastic gradient algorithm which has been widely used for many conventional deep learning models. However, directly applying it to each of our tasks independently can not achieve a joint optimization. The main reason is that our three tasks have significantly different loss functions with highly unbalanced output layers (e.g. 2D, 62D and 1024D). Our Text-CNN is also different from previous MTL model for facial landmark detection, where different tasks are heterogeneous~\cite{Zhang2014}. An important property of our model is that the three tasks have strong hierarchical structure, and can be optimised  sequentially from the low-level region regression to the high-level binary classification. Such hierarchical structure inherently follows basic procedure of our people to identify a text/non-text component. People should first be able to segment a text region arcuately from cluttered background. Then people could make a more confident decision if they recognize the label of a character. Therefore, a reliable high-level decision is strongly built on robust learning of the multi-level prior knowledge.


The training process starts from joint learning of the two auxiliary tasks for text region regression and character recognition. The text region task aims to regress a binary mask of the input image through the deconvolution network. It enables the model with meaningful low-level text information which is important to identify the text and non-text regions at pixel level. Joint optimization of both auxiliary tasks endows the network to learn shared features of multiple characters having a same label, while learning discriminative features between characters of different labels.

\begin{figure*}
\begin{center}
\subfigure[Image]  {\includegraphics[width=4.2cm]{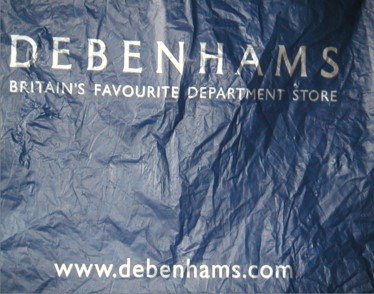}}
\subfigure[CNN]  {\includegraphics[width=4.2cm]{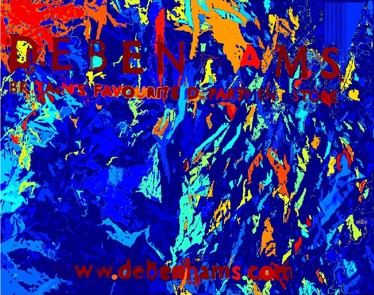}}
\subfigure[CNN-W]  {\includegraphics[width=4.2cm]{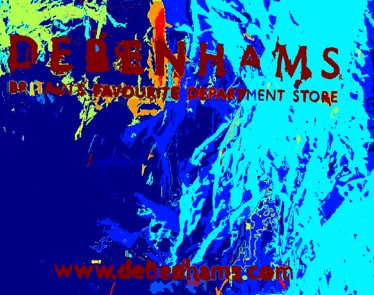}}
\subfigure[Text-CNN]  {\includegraphics[width=4.2cm]{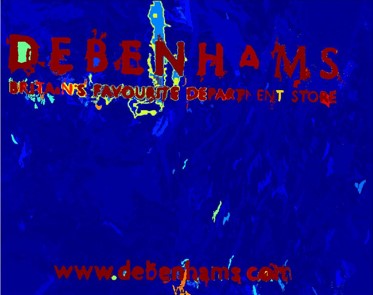}}
\end{center}
   \caption{Confident maps of the conventional CNN, the CNN of \cite{Huang2014,Wang2012} (CNN-W), and Text-CNN.}
\label{fig:confidentmaps}
\end{figure*}

Then the region task is stopped when the main task starts. The main reason for early stoping the regression task is that the low-level knowledge has been embedded into the model successfully after a number of training iterations (e.g. 30,000 iterations in our training process). By jointly learning, the low-level pixel information can be abstracted correctly by the higher-level character label. Still keeping the region task would cause the model overfitting to the low-level task, due to significant deference of their convergence rates. The character task continues with training of the main task until the model is finally optimized.

The label task brings meaningful character-level information (including the low-level character region and its abstracted label) to the main task. Such information may enable the main task with an additional strong feature that discriminates the text and non-text reliably, e.g. a model can identify a component as text confidently, if it knows what exactly this character is. This is crucial to identify some highly confused components, and thus is the key to the robustness and discriminant of our model. Essentially, the label task keeps smooth of the whole training process, by transforming the low-level knowledge to the high-level binary decision. From the viewpoint of parameter optimization, the auxiliary tasks actually work as an important pre-training for our Text-CNN, which has been proven to be importance for seeking a better model optimization \cite{Zeiler2014}.

To demonstrate efficiency of our design, we present the  low-level filters (from the $1$st convolutional layer) learned by three different CNN models in Fig.~\ref{fig:filters}.  The positive impact of each auxiliary task for learning the Text-CNN are shown clearly. Experiment details will be described in Section IV. As shown, the filters learned by the conventional CNN may capture main low-level structures of the character images. However, the filter maps are distorted heavily by an amount of noise, indicating that the binary supervision can not provide informative low-level information for learning robust filters. Obviously, the noise is reduced considerably by using the additional character label task, which provides more detailed character information that facilitates learning of the network. As expected, our Text-CNN with both auxiliary tasks learns a number of high-quality flitters, which not only preserve rich low-level structure of the text, but also are  strongly robust against the background noise.

Robustness and discriminative capability of the Text-CNN are further demonstrated in Fig.~\ref{fig:text-cnn-comp}, where we show (reshaped) feature maps of the last fully-connected layers on a number of challenging samples. As demonstrated by the confident scores  (red bars under the feature maps) and the key discriminative features (located by red circles), the Text-CNN consistently yields high confident scores on these highly ambiguous text components, and generates very low scores for those extremely complicated background components, including many text-like strokes. By contrast, the conventional CNN only works well on clear text components. It has confused scores on the ambiguous ones and complicated background components. The advances of Text-CNN are further demonstrated in the confident maps shown in Fig. \ref{fig:confidentmaps}.

\subsection{Contrast-Enhanced MSERs}
\begin{figure*}
\begin{center}
\subfigure[Contrast Regions (step one)]  {\includegraphics[height=4cm, width=4.2cm]{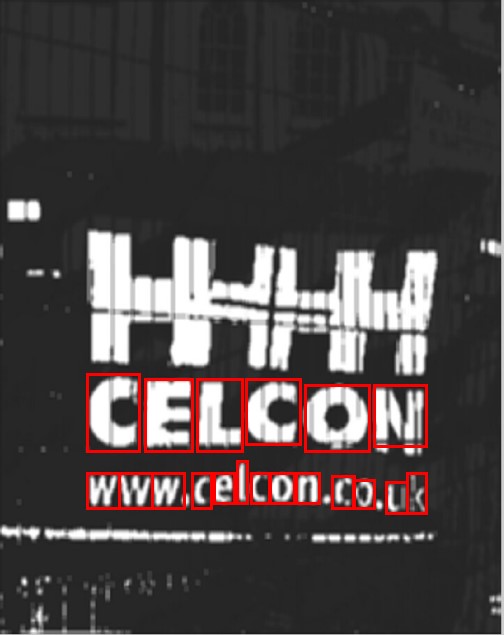}}
\subfigure[Contrast Regions (step two)]  {\includegraphics[height=4cm,width=4.2cm]{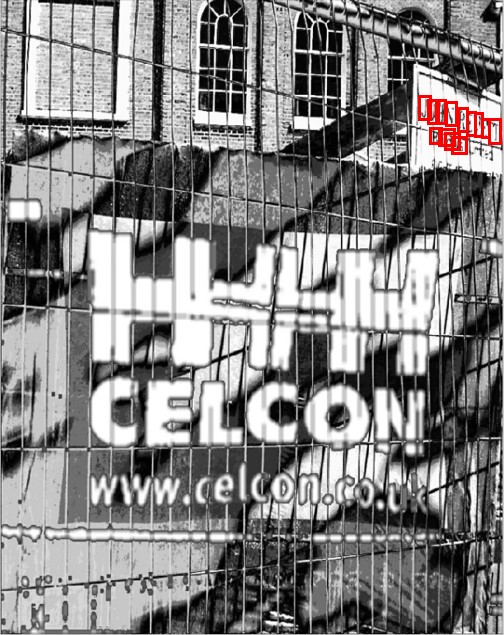}}
\subfigure[CE-MSERs]  {\includegraphics[height=4.2cm,width=4.2cm]{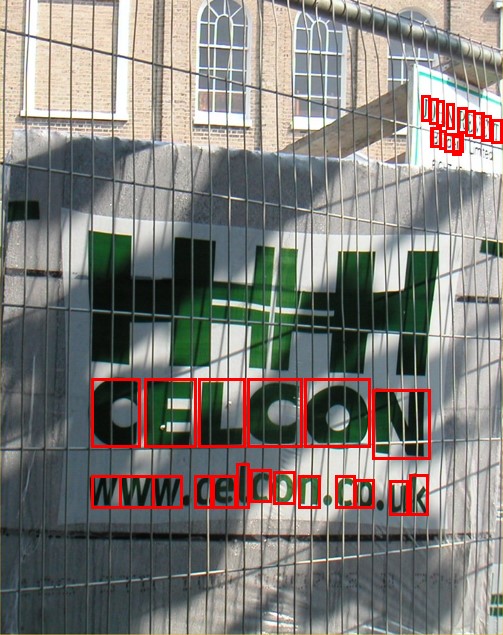}}
\subfigure[MSERs]  {\includegraphics[height=4.2cm,width=4.2cm]{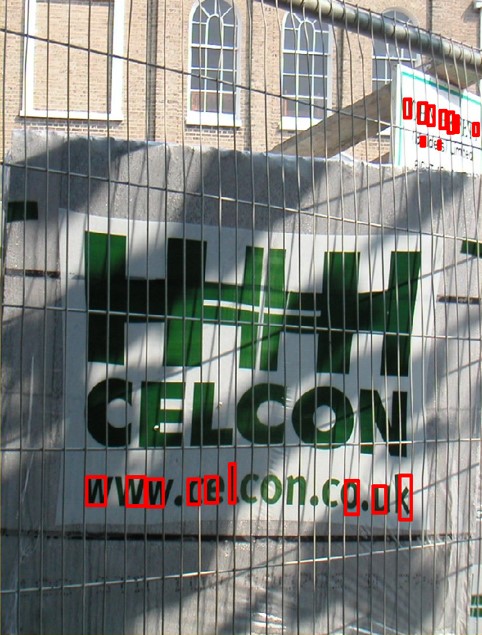}}
\end{center}
   \caption{Contrast-region maps of the CE-MSERs, and comparison of CE-MSERs and MSERs. We discard the bounding boxes which are not matched with the ground truth for a better display.}
\label{fig:cemser}
\end{figure*}

As discussed, the MSERs algorithm has strong ability to detect many challenging text components, by considering each of them as a `stable' extremal region. It has achieved remarkable performance on current text detection systems~\cite{Huang2014,Yin2014}.  However, low-level nature of the MSERs detector largely limits its performance. First, the text components generated by the MSERs are easily distorted  by various complicated background affects, leading to numerous incorrect detections, such as connecting the true texts to the background components, or separating a single character into multiple components. This makes it difficult to identify the true texts in the subsequent filtering step. Second, some text components are ambiguous with low-contrast or low-quality characters. They may not be defined as the `stable' or extremal regions, and thus are discarded arbitrarily by the MSERs. Importantly, it is difficult or impossible to recover these components in the subsequent processes, leading to a significant reduction on recall.

To improve the performance of current MSERs based methods, a crucial issue is to enable it to detect as much true text components as possible. Toward this, we develop an efficient approach that enhances local stability of the text regions, which turns out to mitigate low-level distortions effectively. Thus we wish to enhance region-level contrast of natural images. Motivated from recent methods for salient detections \cite{Fu2013}, we propose a two-step contrast enhancement mechanism.  In the first step, we apply cluster based approach developed in \cite{Cheng2011} for computing the cluster cues globally from the image, e.g. the contrast and spatial cues. Then the contrast region is defined by fusing both cues as \cite{Fu2013}. This approach can strongly enhance the contrast of most dominant or large regions, but it may discard some small-size components as well, as shown in Fig.~\ref{fig:cemser} (a).

To circumvent this problem, we develop the second step to improve contrast of the small-size regions. We compute dominated colors from the remained non-contrast regions. Then the contrast regions are computed in the remained regions by using the color space smoothing method as \cite{Cheng2011}.  The resulted contrast region maps computed by both steps are shown in Fig.~\ref{fig:cemser}. Finally, the original MSERs algorithm is implemented on both contrast region maps and together with the original image, to generate final CE-MSERs components. Details of the CE-MSERs are described in Alg.~\ref{alg_CMSER}.

\begin{algorithm}[H]
  \caption{CE-MSER proposals generation}
  \label{alg_CMSER}
  \begin{algorithmic}[1]

    \STATE \textbf{input}: a natural image
    \STATE Transform the color space from RGB to Lab.
\textbf{    \STATE Step I}
    \STATE Cluster all pixels into $K$ classes using k-means.
    \STATE Compute contrast cue of cluster as \cite{Fu2013}.
    \STATE Compute spatial cue of cluster as  \cite{Fu2013}.
     \STATE Multiply both cues to get the contrast value
     \STATE Generate contrast region map I.
\textbf{    \STATE Step II}
    \STATE Quantize the remaining color from $255^{3}$ to $12^{3}$.
    \STATE Compute color histogram to select the dominant ones.
    \STATE Enhance contrast of similar colors by color space smoothing \cite{Cheng2011}.
    \STATE Generate contrast region maps II.
    \STATE Implement MSERs on the contrast region map I, II and original image.

    \RETURN CE-MSER components.
  \end{algorithmic}
\end{algorithm}

The detection results by the MSERs and CE-MSERs are compared in Fig.~\ref{fig:cemser}. Obviously, the CE-MSERs collect more true text components, some of which are extremely challenging for original MSERs detector. To further demonstrate efficiency of the CE-MSERs, we compute character-level recall (by $\geq0.5$ overlapping) on the ICDAR 2011 dataset. Our CE-MSERs improve the recall of original MSERs from $88.3\%$ to $91.7\%$, which is considerable for current low-level text detectors. This turns out to boost the performance on final detection, with $6\%$ improvement on recall, as shown in Table \ref{tab:component experiments}. More discussions will be presented in Section IV.

The pipeline of full text detection system is as follows. The CE-MSERs detector is first applied to a natural image to generate text components. Then the proposed Text-CNN is adopted to filter out the non-text components. Finally, text-lines are constructed straightforward by following previous work in \cite{Huang2014,Yao2012}. First, two neighboring components are grouped into a pair if they have similar geometric properties, such as horizontal locations, heights and aspect ratios. Second, we merge the pairs sequentially if they include a common component, and at the same time, have similar orientations.  A text-line is constructed until no pair can be merged further. Optionally, a text-line is broken into multiple words based on the horizontal distances of characters or words.

%
%

\section{Experiment and Results}
\begin{figure}
\begin{center}
\includegraphics[width=9cm]{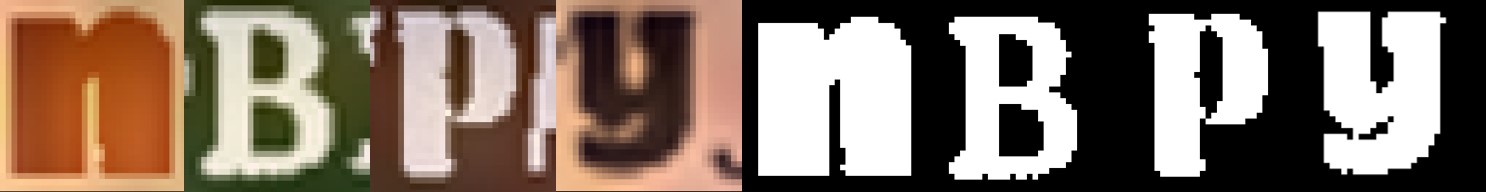}
\end{center}
   \caption{Image samples and their text region masks from the \emph{CharSynthetic}.}
\label{fig:CharSamples}
\end{figure}

We evaluate efficiency of the proposed Text-CNN and CE-MSERs individually. Then our system is tested on four benchmarks for scene text detection: the ICDAR 2005 \cite{Lucas2005}, ICDAR 2011 \cite{Shahab2011}, ICDAR 2013 \cite{Karatzas2013} and MSRA-TD500 \cite{Yao2012} databases.

\subsection{Experimental Setup and Datasets}
\begin{figure}
\begin{center}
\includegraphics[width=9cm]{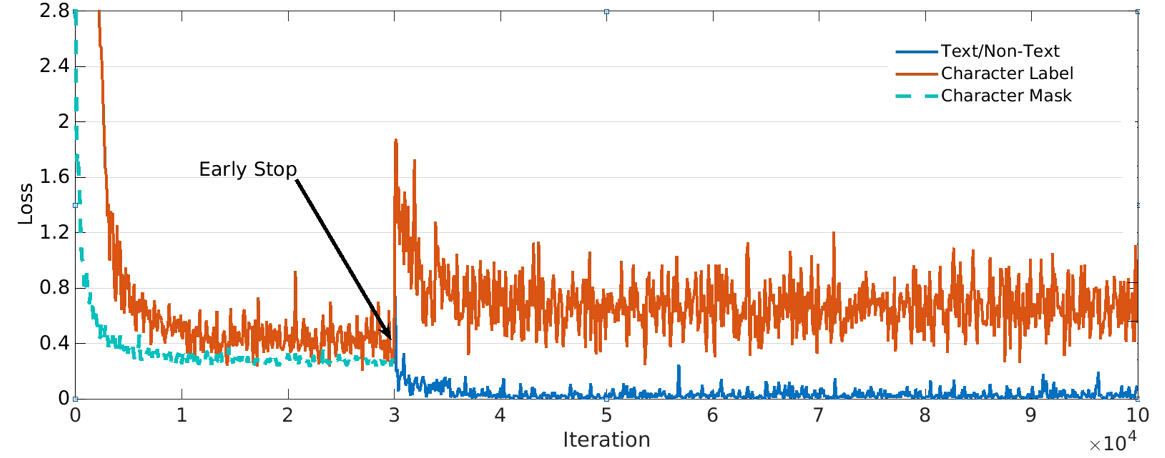}
\end{center}
   \caption{Training loss of the Text-CNN. The character label task runs through the whole training process. The training data is changed from the $CharSynthetic$ to the $charTrain$ after the early stop at 30K iterations.}
\label{fig:trainLoss}
\end{figure}

The Text-CNN is trained on character-level by using two datasets, \emph{CharSynthetic} and \emph{CharTrain}. In the \emph{CharSynthetic}, we synthetically generated 80,141 character images, each of which has both character mask and its label. Some examples are shown in Fig.~\ref{fig:CharSamples}. This dataset is applied for jointly training the low-level text region and character label tasks. The \emph{CharTrain} includes 17,733 text (character) and 85,581 non-text samples, which are cropped from training set of the ICDAR 2011 dataset \cite{Shahab2011}. Each sample has binary text/non-text information and 62-class character label, which are used to joint train the character label task and the main task.
%
%
The training of Text-CNN starts from joint learning of mask regression and character recognition tasks, by using the \emph{CharSynthetic}. The loss parameters are manually set to $\lambda_1=1$ and $\lambda_2=0.3$. The mask regression task is stopped at 30K iterations. Then the main task starts, together with the continued character label task, whose loss parameter is then changed to $\lambda_1=0.3$.
The tasks are trained with further 70K iterations by using the \emph{CharTrain}. The curves of multiple training losses are presented in Fig.~\ref{fig:trainLoss}, where they converge stably at 30K and 100K iterations.

\begin{figure}
\begin{center}
\includegraphics[width=8cm]{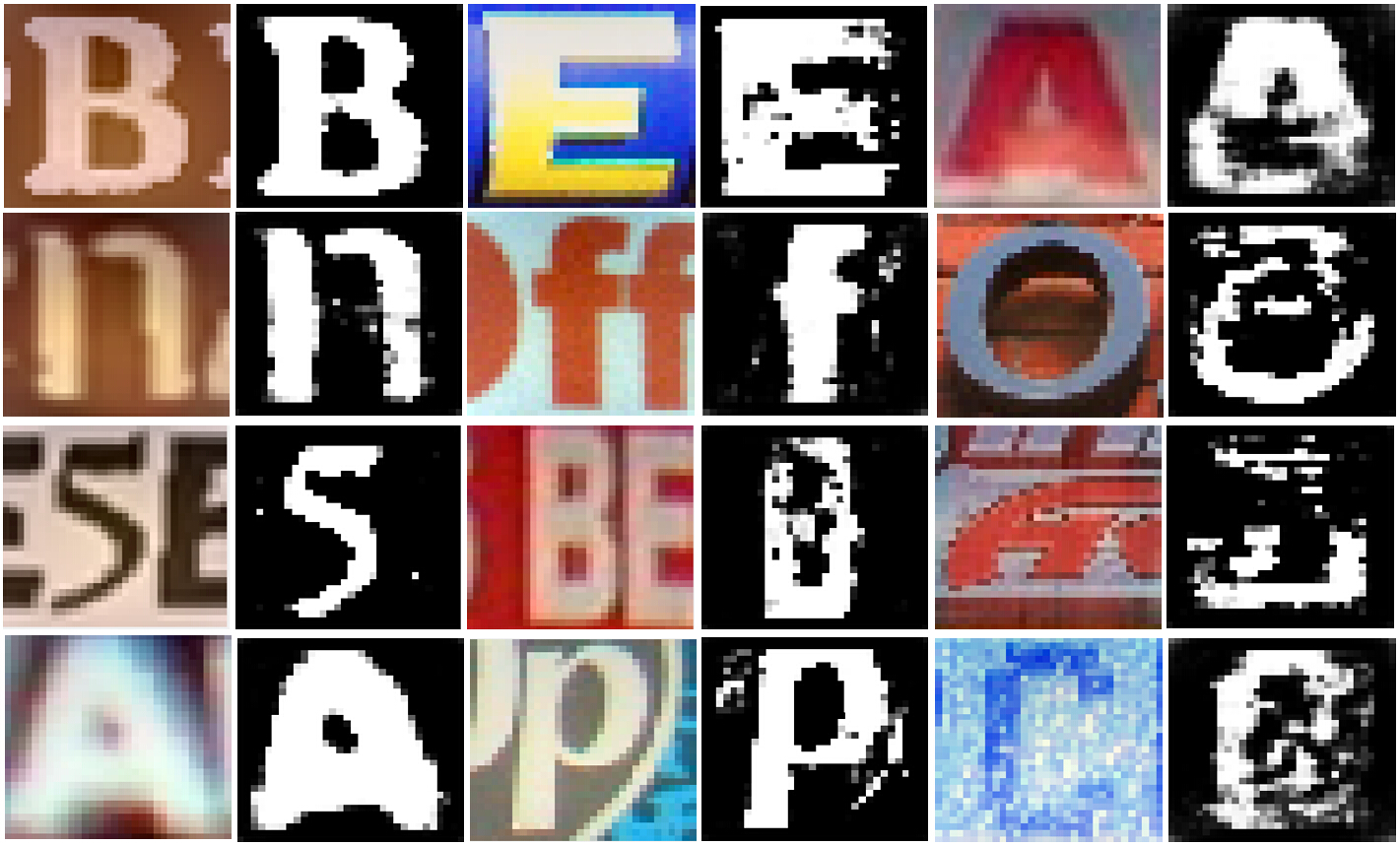}
\end{center}
   \caption{Examples of the estimated masks.}
\label{fig:regressExample}
\end{figure}

We investigate the performance of Text-CNN with various values of $\lambda_1$ and $\lambda_2$ in Table \ref{tab:lamda}. The experiments were conducted as follow.  In stage one (first 30K iterations), we fixed $\lambda_1=1$ and changed the value of $\lambda_2$. We computed 62-class character recognition accuracy ($P_{charLabel}$), and L2 distance ($D_{Mask}$) between the estimated mask and ground true mask on 20,036 synthetically generated character images. Examples of the estimated masks are shown in Fig. \ref{fig:regressExample}. In stage two, we computed the 62-class and text/non-text classification accuracies on the $testChar$, by changing the value of $\lambda_1$. As can be found, the model can not converge well when the value of $\lambda_2$ is larger than 0.5 in the stage one, while the impact of $\lambda_1$ is much less significance in the stage two.


%
%
%
%

\begin{table}[tb]
\caption{We evaluate parameters of the Text-CNN: $\lambda_1$ and $\lambda_2$. In the Stage one (first 30K iterations), we fix $\lambda_1=1$ and change the value of $\lambda_2$.  In the Stage two, we very the value of $\lambda_1$.}

\begin{tabular}{l|c|c|c|c|c}
\hline
$\lambda_2$ (Stage 1) &0.9 &0.7 &0.5 &0.3 &0.1  \\\hline
$P_{charLabel}$ &1.6\% &1.8\% &6.4\% &84.5\% &84.3\% \\\hline
$D_{mask}$ &147.5 &141.7 &131.8 &19.8 &21.1 \\\hline
\hline
$\lambda_1$ (Stage 2) &0.9 &0.7 &0.5 &0.3 &0.1  \\\hline
$P_{charLabel}$ &87.5\% &87.3\% &87.4\% &87.5\% &86.1\%  \\\hline
$P_{Text/Non-Text}$ &92.6\% &92.3\% &92.6\% &92.6\% &92.3\%  \\\hline
\end{tabular}\label{tab:lamda}
\end{table}


For testing the Text-CNN, we collect a character-level dataset: the \emph{CharTest}, which includes 5,751 character images and 11,502 non-text components. The character images are cropped manually from the ICDAR 2011 test set, while the non-text ones are randomly selected from the image background of same dataset.

\begin{figure}
\begin{center}
\subfigure[Enlarged CE-MSER component]{\includegraphics[height=2cm,width=8cm]{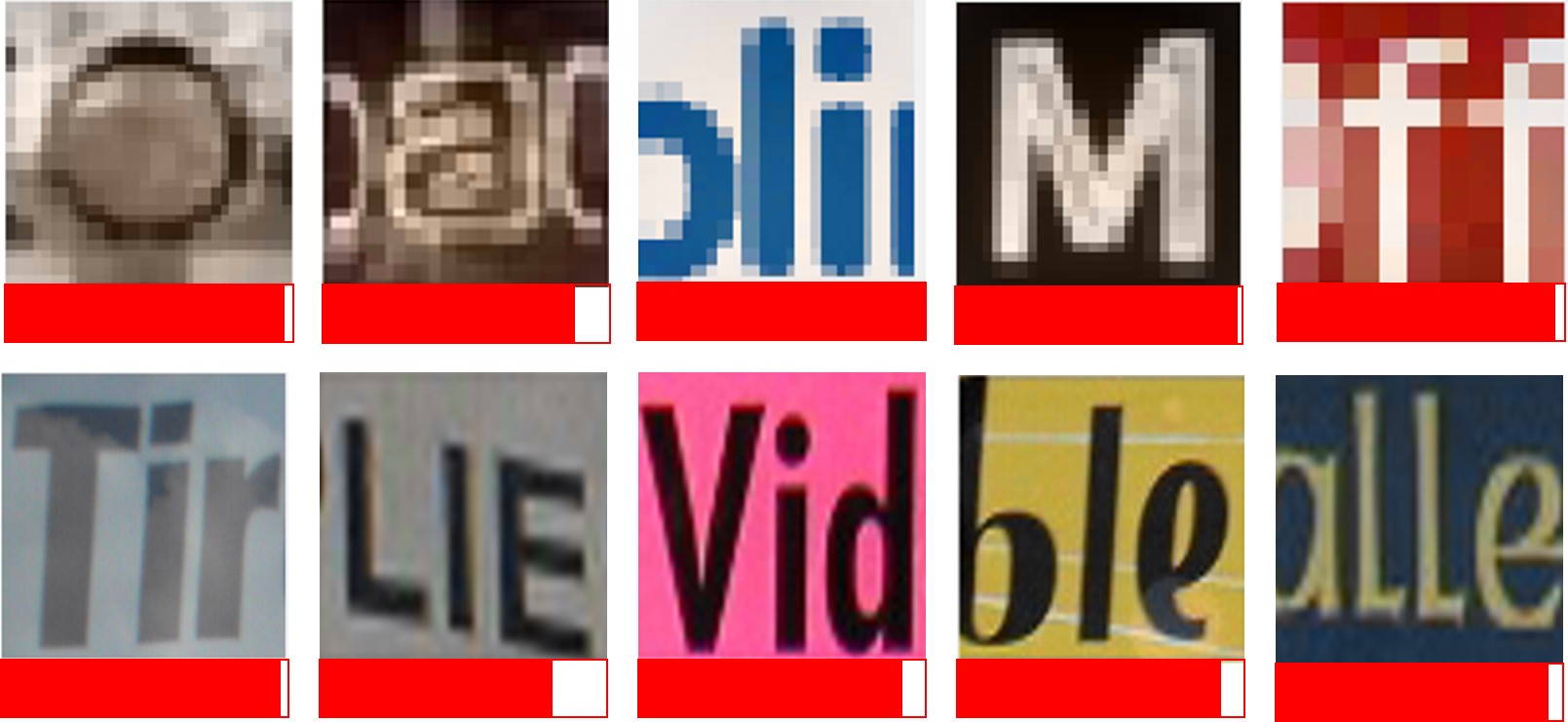}}
\subfigure[Correct recognition examples only by CNN with mask regression]{\includegraphics[height=2cm,width=8cm]{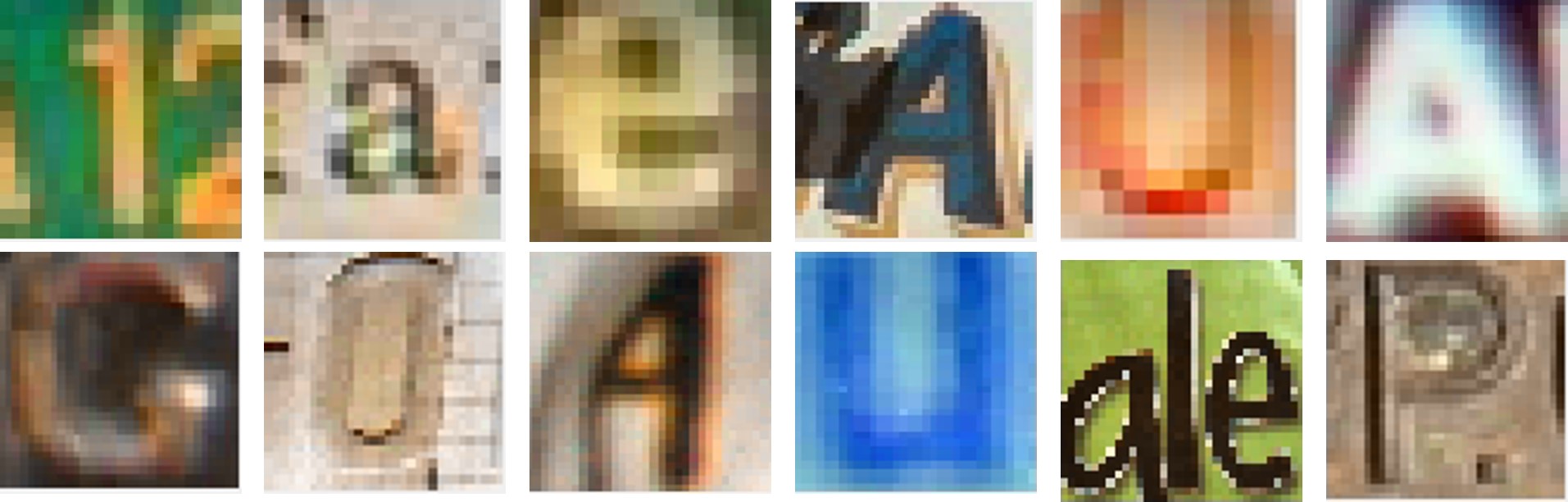}}
\end{center}
   \caption{(a) Examples of enlarged CE-MSER component by its long side, with corresponding Text-CNN score. (b) Some component examples for character recognition, which are all classified correctly by the CNN with mask regression, but are all failed without the mask regression.}
\label{fig:multiChar}
\end{figure}

As mentioned, the input of Text-CNN is a 32$\times$32 patch. We do not directly resize the detected component, instead we enlarge each CE-MSER region to a square area by using its long side, as shown in Fig. \ref{fig:multiChar} (a). Therefore, each input patch may include neighboring characters or parts of them, allowing the Text-CNN to classify it more reliably by exploring surrounding context. This is particularly important to those components which are easily confused with cluttered background, such as `I', `1', `l', etc. This operation is implemented on both training and test data. On the other hand, the original CE-MSER or MSER components may include multiple characters, while a single character may also be detected by multiple components due to tree structure of the MSER. The Text-CNN aims to retain all possible components containing true characters. It is able to classify a multi-character component as a positive detection, as shown in Fig. \ref{fig:multiChar} (a). In general, the confident score decreases when a component has more than three characters.

Our system is evaluated on the ICDAR 2005, 2011 and 2013 Robust Reading Competition benchmarks and the MSRA-TD500 database. The ICDAR 2005 \cite{Lucas2005} has 258 images in the training set, and 251 images for testing with size varying from 307 $\times$ 97 to 1280 $\times$ 960. The ICDAR 2011 \cite{Shahab2011} includes 229 and 255 images for training and testing, respectively. There are 299 training images and 233 test images in the ICDAR 2013 \cite{Karatzas2013}. The MSRA-TD500 database \cite{Yao2012} has 500 images containing multi-orientation text lines in different languages (e.g. Chinese, English or mixture of both). The training set contains 300 images, and the rest is used for testing.

\subsection{Results and Comparisons}
\begin{figure*}
\begin{center}
\subfigure[CNN]  {\includegraphics[height=3.5cm,width=4.2cm]{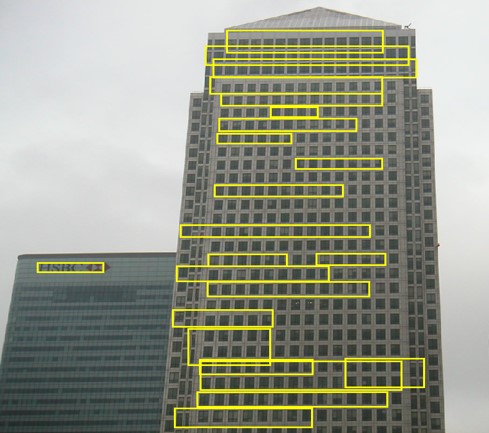}}
\subfigure[CNN-W]  {\includegraphics[height=3.5cm,width=4.2cm]{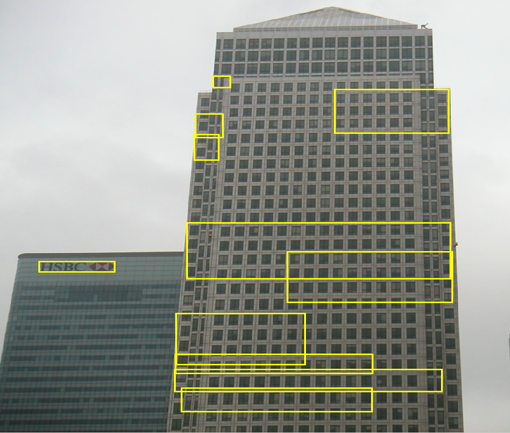}}
\subfigure[Text-CNNL]  {\includegraphics[height=3.5cm,width=4.2cm]{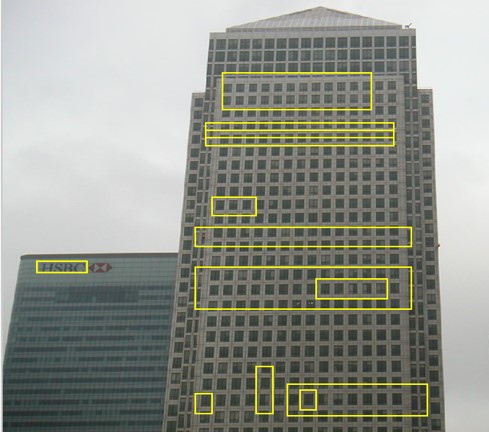}}
\subfigure[Text-CNN]  {\includegraphics[height=3.5cm,width=4.2cm]{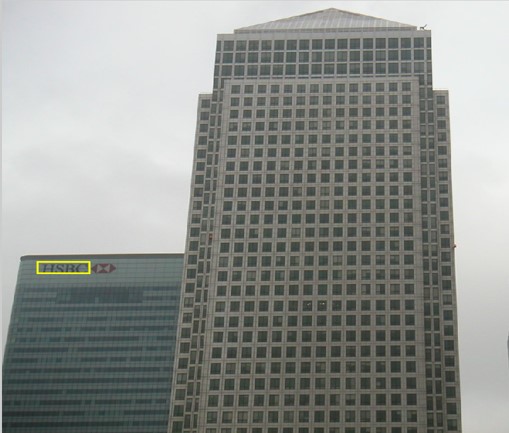}}
\end{center}
   \caption{Detection results of the conventional CNN, CNN of \cite{Huang2014,Wang2012} (CNN-W) and Text CNN, by using the same CE-MSERs detector.}
\label{fig:building}
\end{figure*}

To investigate the efficiency of each proposed component, we first evaluate the Text-CNN on text component classification. Then the contribution of individual Text-CNN or CE-MSERs to our final detection system is investigated. Finally, the performance of final system are compared extensively with recent results on four benchmarks.

\subsubsection{Text-CNN on text component classification}

The experiments were conducted on the \emph{CharTest}, and results are presented in Table~\ref{tab:CharTest}. To demonstrate the impact of different types of supervisions, we conducted several experiments which were trained and tested on a same dataset for fair comparisons. The baseline is conventional CNN trained by just using text/non-text information. As can be seen, the Text-CNN outperforms the conventional CNN substantially, reducing error rate from $9.8\%$ to $6.7\%$ by using informative low-level supervision.
The individual character mask and character label information improve the CNN by 1.4\% and 2.0\%, respectively.

We further investigate their error rates on separated character and non-character subsets of the \emph{CharTest}. As can be found in Table ~\ref{tab:CharTest}, the main errors are raised in the character set by classifying true characters into non-text components. There were two observations from the results. First, the character label information is of great importance for improving the accuracies in the character set, whose error rate is reduced from 28.3\% to 17.5\%. It may be due to that some ambiguous character structures are easily confused with text-like background objects (e.g. bricks, windows and
leaves) which often have similar low-level mask structures as true characters or character strokes. The mid-level character label information is indeed helpful to discriminate them by telling the model what stroke structures of the true characters are. This turns out to improve the final recall. Second, the mask information helps to reduce the error in classifying background components into the characters, resulting in a reduction of false alarms.  We also compare our model against the CNN classifier used in \cite{Huang2014,Wang2012} (referred as CNN-W). The performance of Text-CNN is also over $8.6\%$ of the CNN-W trained purely on text/non-text supervision. The strong ability of the Text-CNN is further demonstrated in Fig.~\ref{fig:building}, where it shows strong robustness against a large amount of building windows. These windows have similar low-level properties as the text patterns, and the character mask and label information are greatly helpful to discriminate them, which turns out to yield a high precision.

In addition, we evaluate the impact of mask regression on character recognition task. The experiments were conducted on the \emph{CharTest}, with 5,751 character images of 62 classes. We compare performance of the conventional CNN and mask pre-trained CNN, which achieve 14.2\% and 12.7\% error rates respectively. Obviously, the mask regression also improves the performance of CNN on character recognition. Some challenging examples which are only recognized successfully by the CNN with mask regression are presented in Fig. \ref{fig:multiChar} (b).  This further confirms the efficiency of mask regression for pre-training the CNN model.

\begin{table}[tb]
\centering \caption{Independent impact of character mask or character label for text and non-text component classification and recognition. Experiments were conducted on ICDAR 2011 test set with character-level annotation (the $CharTest$ set).}
\begin{tabular}{l|c|c|c}
\hline
Method          &CharTest      &Text (5751)  &Non-Text (11550) \\\hline
\hline
\multicolumn{4}{c} {Text$/$Non-Text Classification (Error Rate)}\\\hline
CNN-W           &8.6\%         &--                &-- \\\hline
CNN             &9.8\%         &28.3\%            &2.2\%\\\hline
CNN-Mask        &8.4\%         &24.7\%            &1.8\%\\\hline
CNN-Label       &7.8\%         &17.5\%            &4.0\% \\\hline
Text-CNN	    &6.7\% 	       &18.6\%	          &1.8\% \\\hline
\hline
\multicolumn{4}{c} {62-Class Character Recognition}\\\hline
CNN         & --     & 14.2\%& --\\\hline
CNN-Mask  	& --	 & 12.7\%& --	\\\hline
\end{tabular}\label{tab:CharTest}
\end{table}

\begin{table}[tb]
\centering \caption{Independent contribution of the Text-CNN and CE-MSERs (on the ICDAR 2011 dataset).}
\begin{tabular}{l|c|c|c}

\hline

Method                                    &$Precision$         &$Recall$    &$Fmeasure$   \\\hline \hline
& \multicolumn{3}{|c}{with Text-CNN}\\\hline
MSERs                            &0.89     &0.68     &0.78  \\\hline
CE-MSERs                         &0.91      &0.74    &0.82  \\\hline\hline

& \multicolumn{3}{|c}{with CE-MSERs}\\\hline

CNN                             &0.85      &0.71    &0.77  \\\hline
Text-CNNL                      &0.88      &0.72    &0.79  \\\hline
Text-CNN                        &0.91      &0.74    &0.82  \\\hline
\end{tabular}\label{tab:component experiments}
\end{table}

\subsubsection{Evaluation on the individual  CE-MSERs or Text CNN}

We conducted experiments on the ICDAR 2011,
and experimental results are presented in Table \ref{tab:component experiments}. As discussed, our CE-MSERs detector improves the original MSERs with an about $3\%$ character-level recall. This leads to a surprising improvement on the recall of final detection (from $68\%$ to $74\%$). The large improvement indicates that the CE-MSERs is able to detect some challenging text patterns, which are of importance for detecting or retaining the whole text-lines or words missed by the original MESRs based methods. For example, a word can be easily broken into two separate words in text-line construction, if a character is missed in the middle location. The broken words are considered as false detections, reducing both precision and recall. This can be verified by the improved precision by the CE-MSERs. Both improvements lead to a $4\%$ improvement on the F-measure.

The advantages of Text-CNN are demonstrated on the remarkable improvement on precision. The Text-CNN improves conventional CNN substantially from $85\%$ to $91\%$, where the single character task obtains a $3\%$ improvement. Similarly, strong capability of the Text-CNN also results in an increase of recall, since a word is not correctly detected if it includes the background components which are not filtered out robustly.

The running time of the CE-MSER and Text-CNN was evaluated on the ICDAR 2011. The average time for the CE-MSER is about 4.1s per images by our MATLAB implementation, compared to 1.2s of the original MSER (also implemented in MATLAB). The average time for the Text-CNN is about 0.5s per image implementing in CAFFE framework \cite{Jia2014} with a single GPU. The implementation of CE-MSER can be accelerated significantly with more engineering optimization, e.g. about 0.3s/image by the MSER, as reported in \cite{Busta2015}. 

\subsubsection{Evaluation on full text detection}

\begin{table}[tb]
\centering \caption{Experimental results on the ICDAR 2005 dataset.}
\begin{tabular}{l|c|c|c|c}

\hline

Method                                      &Year    &$Precision$         &$Recall$    &$Fmeasure$\\\hline \hline
\textbf{Our method}                         & --     &\textbf{0.87}       &\textbf{0.73}    &\textbf{0.79}     \\\hline\hline
MSERs-CNN \cite{Huang2014}                          & 2014   &0.84       &0.67    &0.75 \\\hline
SFT-TCD~\cite{Huang2013}                    & 2013   &0.81       &0.74    &0.72 \\\hline
Yao \emph{et al}.~\cite{Yao2012}            & 2012   &0.69       &0.66    &0.67  \\\hline
Chen \emph{et al}.~\cite{Chen2012}          & 2012   &0.73       &0.60    &0.66  \\\hline
Epshtein \emph{et al}.~\cite{Epshtein2010}  & 2010   &0.73       &0.60    &0.66  \\\hline
Yi and Tian~\cite{Yi2013}                   & 2013   &0.71       &0.62    &0.63  \\\hline
Neumann $\&$ Matas~\cite{Neumann2011}        & 2011   &0.65       &0.64    &0.63  \\\hline

Zhang $\&$ Kasturi~\cite{Zhang2010}          & 2010   &0.73       &0.62    &--    \\\hline
Yi $\&$ Tian~\cite{Yi2011}                   & 2011   &0.71       &0.62    &0.62  \\\hline

\end{tabular}\label{tab:results_ICDAR2005}
\end{table}

\begin{table}[tb]
\centering \caption{Experimental results on the ICDAR 2011 dataset.}
\begin{tabular}{l|c|c|c|c}

\hline

Method                                                &Year    &$Precision$         &$Recall$    &$Fmeasure$   \\\hline \hline
\textbf{Our method}                              & --
&\textbf{0.91}       &0.74    &\textbf{0.82}  \\\hline\hline
Zhang  \emph{et al}.\cite{Zhang2015}                 & 2015     &0.84       &\textbf{0.76}    &0.80  \\\hline
MSERs-CNN \cite{Huang2014}                           & 2014     &0.88       &0.71    &0.78  \\\hline
Yin \emph{et al}.~\cite{Yin2014}                      & 2014   &0.86       &0.68    &0.76 \\\hline
Neumann $\&$ Matas~\cite{Neumann2014}                 & 2013   &0.85        &0.68    &0.75 \\\hline
SFT-TCD~\cite{Huang2013}                              & 2013   &0.82       &0.75    &0.73 \\\hline
Neumann $\&$ Matas~\cite{Neumann2013}                  & 2013   &0.79       &0.66    &0.72 \\\hline
Shi \emph{et al}.~\cite{Shi2013}                      & 2013   &0.83       &0.63    &0.72 \\\hline
Neumann $\&$ Matas~\cite{Neumann2012}                  & 2012   &0.73       &0.65    &0.69  \\\hline
Gonz$\acute{a}$lez \emph{et al}.~\cite{Gonzalez2012}  & 2012   &0.73       &0.56    &0.63  \\\hline
Yi $\&$ Tian~\cite{Yi2011}                             & 2011   &0.67       &0.58    &0.62  \\\hline
\end{tabular}\label{tab:results_ICDAR2011}
\end{table}

\begin{table}[tb]
\centering \caption{Experimental results on the ICDAR 2013 dataset. Our performance is compared to the last published results in~\cite{Zhang2015}. }
\begin{tabular}{l|c|c|c|c}

\hline

Method             &Year    &$Precision$         &$Recall$    &$Fmeasure$   \\\hline \hline
\textbf{Our method} & --  &\textbf{0.93}   &0.73    &\textbf{0.82}  \\\hline\hline
Zhang \emph{et al}.\cite{Zhang2015} & 2015   &0.88  &\textbf{0.74}    &0.80  \\\hline
iwrr2014 ~\cite{Zamberletti2014} & 2014   &0.86       &0.70    &0.77 \\\hline
USTB TexStar~\cite{Yin2014}  & 2014   &0.88       &0.66    &0.76 \\\hline
Text Spotter~\cite{Neumann2012}   & 2012   &0.88   &0.65    &0.75 \\\hline
\end{tabular}\label{tab:results_ICDAR2013}
\end{table}


\begin{table}[tb]
\centering \caption{Experimental results on the MSRA-TD500 dataset.}
\begin{tabular}{l|c|c|c|c}

\hline

Method                                      &Year    &$Precision$         &$Recall$    &$Fmeasure$\\\hline \hline

\multicolumn{5}{c}{MSRA-TD500}
\\\hline
\textbf{Our method}                         & --     &0.76       &0.61    &0.69     \\\hline
Yin \emph{et al}.~\cite{Yin2015}            & 2015   &\textbf{0.81}       &\textbf{0.63}    &\textbf{0.71} \\\hline
Yin \emph{et al}.~\cite{Yin2014}            & 2014   &0.71       &0.61    &0.66 \\\hline
Kang \emph{et al}.~\cite{Kang2014}      & 2014   &0.71       &0.62    &0.66    \\\hline
Yao \emph{et al}.~\cite{Yao2014}            & 2014   &0.62       &0.64    &0.61  \\\hline
Yao \emph{et al}.~\cite{Yao2012}            & 2012   &0.63       &0.63    &0.60  \\\hline
\hline

\multicolumn{5}{c}{ICDAR2011}
\\\hline

\textbf{Our method}    & --     &\textbf{0.91}         &\textbf{0.74}     &\textbf{0.82}       \\\hline
Yin \emph{et al}.~\cite{Yin2015}     &2015  &0.84        &0.66        &0.74 \\\hline
\hline

\multicolumn{5}{c}{ICDAR2013}
\\\hline
\textbf{Our method}   & --    &\textbf{0.93}       &\textbf{0.73}      &\textbf{0.82}     \\\hline
Yin \emph{et al}.~\cite{Yin2015}    &2015   &0.84        &0.65        &0.73 \\\hline 			

\end{tabular}\label{tab:results_MSRA}
\end{table}

\begin{figure*}
\subfigure[Successful examples]{\includegraphics[height=10cm,width=13.5cm]{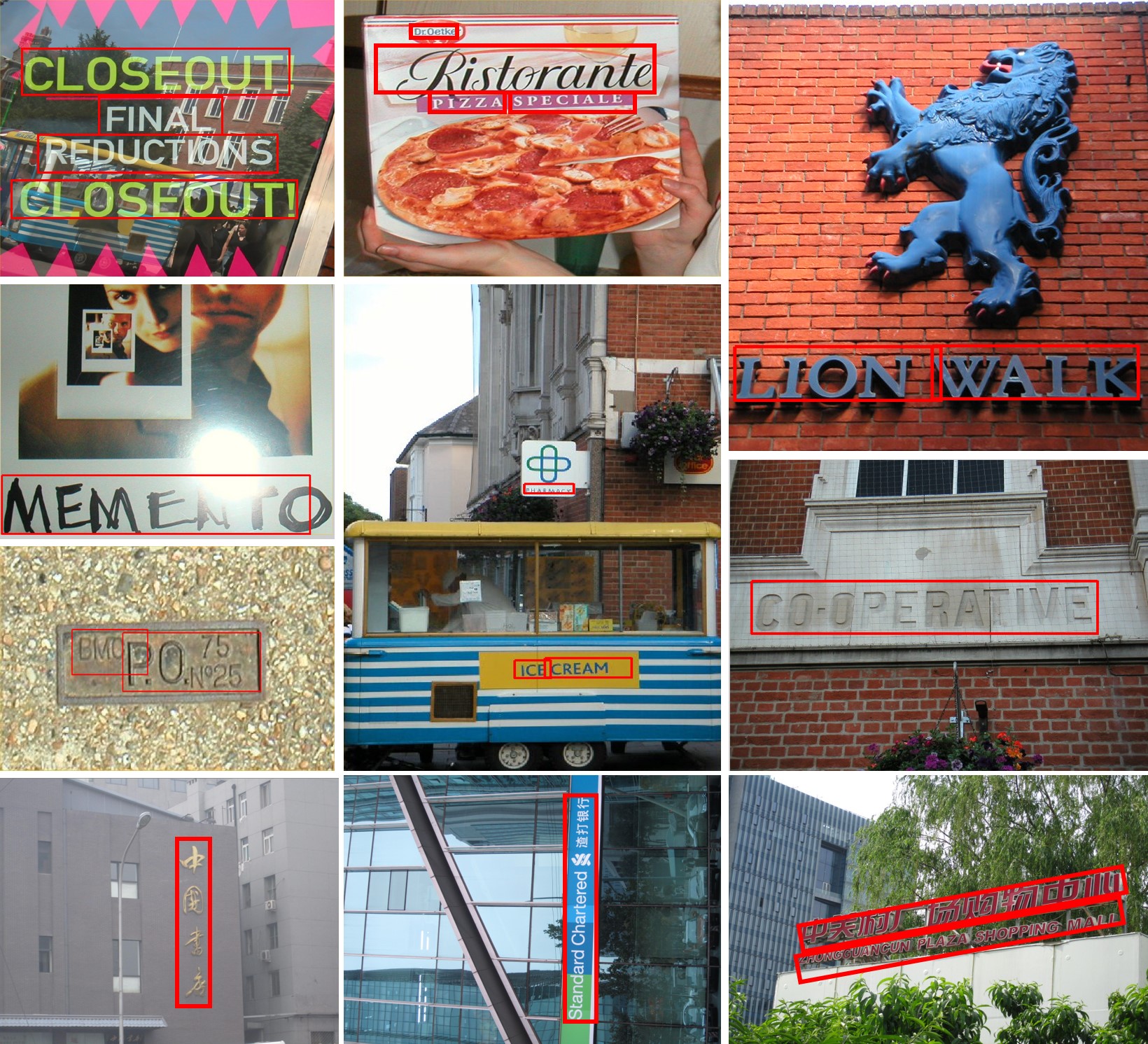}}
\enskip
\subfigure[Failure cases]{\includegraphics[height=10cm,width=4cm]{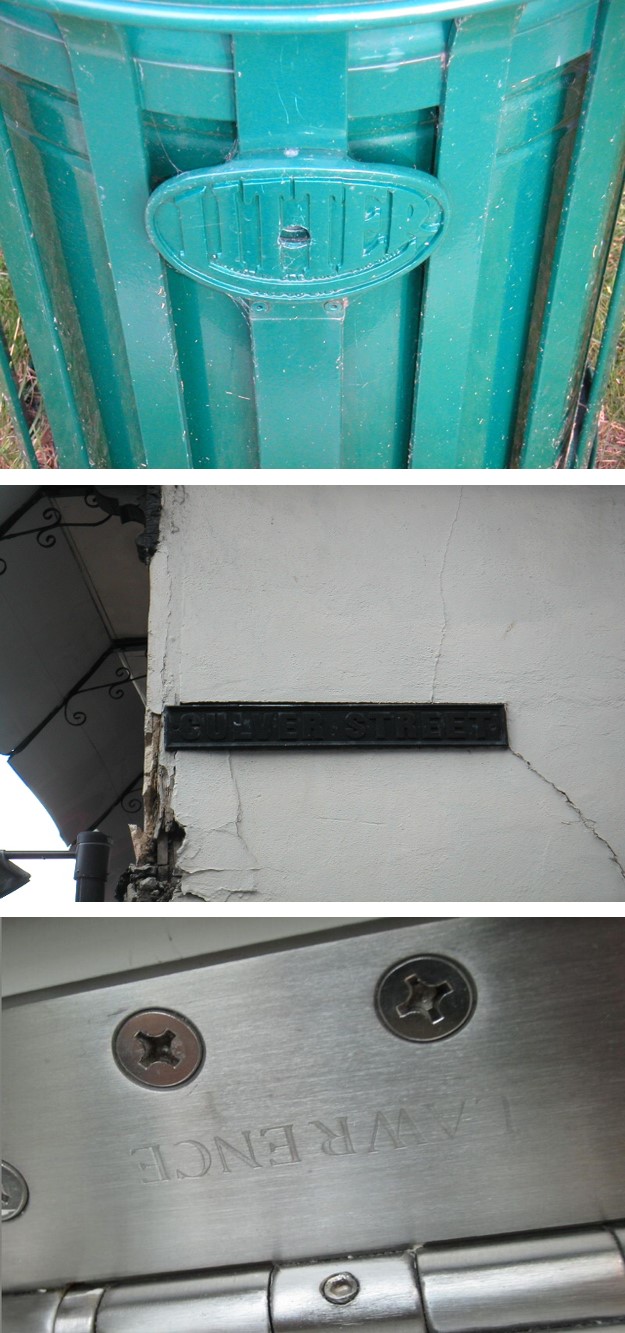}}
   \caption{Successful detection results on changeling examples, and failure cases.}
\label{fig:inclined_textlines.jpg}
\end{figure*}

 The full evaluation of the proposed system was conducted on three benchmarks. We follow standard evaluation protocol of the ICDAR 2011 by using the DetEval tool offered by authors of \cite{Wolf2006}, and the ICDAR 2013 standard in \cite{ICDAR2013}. The performance of the proposed approach in terms of \emph{Precision}, \emph{Recall} and \emph{F-measure}, is compared in Table~\ref{tab:results_ICDAR2005}, \ref{tab:results_ICDAR2011} and \ref{tab:results_ICDAR2013}. Our detection system achieves promising results on all three datasets, which improve recent results considerably.
In the ICDAR 2005 and 2011 datasets, the proposed approach outperforms the MSERs-CNN \cite{Huang2014} substantially with $4\%$ improvement in F-measure. The large improvement on recall is mainly beneficial from strong ability of the enhanced CE-MSERs detector which robustly identifies more distorted text patterns. While the Text-CNN improves discriminative power, so that it reduces the number of false alarms dramatically, and thus boosts the precision. Comparing to the closest performance achieved by Zhang \emph{et al.} \cite{Zhang2015} on the ICDAR 2011,  the Text-CNN obtains a 7\% improvement on precision. Finally, the proposed system is further evaluated on the ICDAR 2013 dataset. It achieves a high precision of 0.93 with 0.73 recall and 0.82 F-measure, which  advance recent published results \cite{Zhang2015}.


The detection results on a number of challenging images are shown in Fig.~\ref{fig:inclined_textlines.jpg}. The correct detection samples demonstrate high performance of our system with strong robustness against multiple text variations and significantly cluttered background.
The failure cases have extremely ambiguous text information and are easily confused with its background. They are even hard to human detection.

\subsubsection{Evaluation on multi-orientation and multi-language text}
The generality of the proposed method to multi-language and multi-orientation text lines is further investigated. We conducted experiments on the MSRA-TD500 dataset. For training the Text-CNN, we synthetically generate 30,000 Chinese character images and corresponding masks. There are totally 2,300 character classes including 2,238 Chinese character classes and 62 English characters in the $CharTrain$. The training process follows previous experimental settings.

The results are reported in Table \ref{tab:results_MSRA}, with comparisons against state-of-the-art performance on this benchmark \cite{Yin2015, Yao2014,Kang2014}. As can be found, our method achieves a F-measure of 0.69, significantly outperforming recent methods of \cite{Yao2014} (with 0.61) and \cite{Kang2014} (with 0.66). Our result comes tantalizingly close to the best performance of 0.71 F-measure achieved by Yin, \emph{et al}. \cite{Yin2015}. Notice that all three methods \cite{Yin2015,Yao2014,Kang2014} compared are specially designed for multi-oriented text detection. On the other hand, we also compare our performance with Yin \emph{et al.}'s  approach \cite{Yin2015} on near-horizontal text in the Table \ref{tab:results_MSRA}, where our method obtains large improvements, by 8-9\% on the ICDAR 2011 and 2013. These results demonstrate the efficiency and generality of our method convincingly. In fact, we aims to solve fundamental challenge for this task by providing more accurate or reliable character candidates. It could be readily incorporated with a more sophisticated method for text line construction (e.g., \cite{Yin2015,Kang2014}), and then better performance can be expected.

\section{Conclusion}

We have presented a new system for scene text detection, by introducing a novel Text-CNN classifier and a newly-developed CE-MSERs detector.
The Text-CNN is designed to compute discriminative text features from an image component.  It leverages highly-supervised text information, including text region mask, character class, and binary text/non-text information. We formulate the training of Text-CNN as a multi-task learning problem that effectively incorporates interactions of multi-level supervision. We show that the informative multi-level supervision are of particularly importance for learning a powerful Text-CNN which is able to robustly discriminate ambiguous text from complicated background. In addition, we improve current MSERs by developing a contrast enhancement mechanism that enhances region stability of text patterns. Extensive experimental results show that our system has achieved the state-of-the-art performance on a number of benchmarks.

\bibliographystyle{IEEEtran}
\bibliography{egbib2}

\begin{IEEEbiography}[{\includegraphics[width=1in,height=1.25in,clip,keepaspectratio]{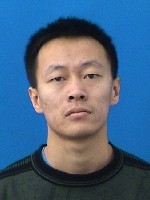}}]{Tong He}
  received B.S. degree in Marine Technology from Tianjin University of Science and Technology, Tianjin, China, in 2013, where he is currently pursuing the M.S. degree with School of Remote Sensing and Information Engineering, Wuhan University.  He has been with Shenzhen Institutes of Advanced Technology Chinese Academy of Sciences, where he is a visiting graduate students. His research interests are text detection, text recognition, general object detection and scene classification.
\end{IEEEbiography}

\begin{IEEEbiography}[{\includegraphics[width=1in,height=1.25in,clip,keepaspectratio]{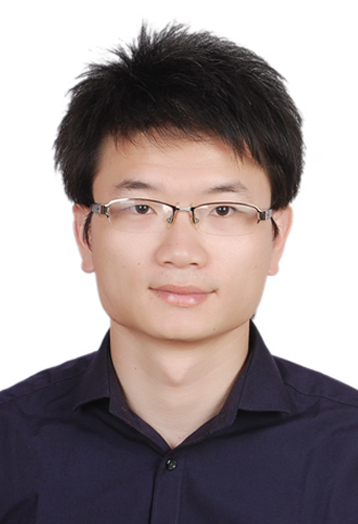}}]{Weilin Huang}
(M'13)  received PhD degree in electronic engineering from the University of Manchester (UK) in December 2012. He got his BSc in computer science and MSc in internet computing from the University of Shandong (China) and University of Surrey (UK), respectively. Currently, he is working as a Research Assistant Professor at Chinese Academy of Science, and a joint member in the Multimedia Laboratory, Chinese University of Hong Kong. His research interests include computer vision, machine learning and pattern recognition. He has served as reviewers for several journals, such as IEEE Transactions on Image Processing, IEEE Transactions on Systems, Man, and Cybernetics
(SMC)-Part B and Pattern Recognition. He is a member of IEEE.
\end{IEEEbiography}

\begin{IEEEbiography}[{\includegraphics[width=1in,height=1.25in,clip,keepaspectratio]{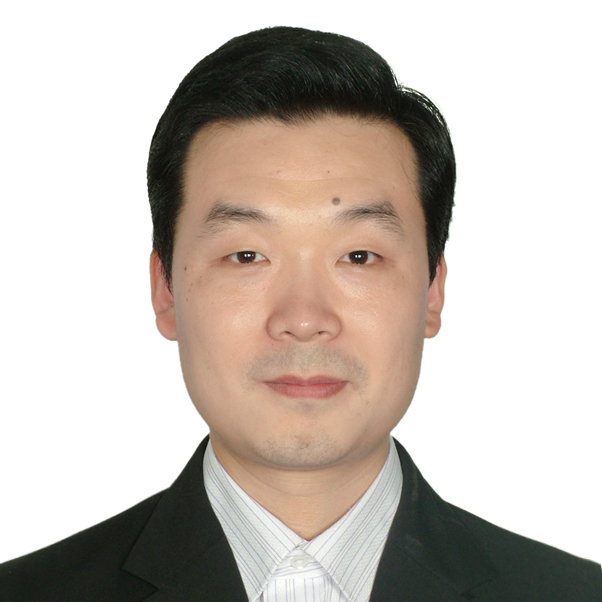}}]{Yu Qiao}
(SM'13) received the Ph.D. degree from the University of Electro-Communications, Japan, in 2006. He was a JSPS Fellow and Project Assistant Professor with the Unversity of Tokyo from 2007 to 2010. He is currently a Professor with the Shenzhen Institutes of Advanced Technology, Chinese Academy of Sciences. His research interests include pattern recognition, computer vision, multi-media, image processing, and machine learning. He has publised more than 90 papers. He received the Lu Jiaxi Young Researcher Award from the Chinese Academy of Sciences in 2012.
\end{IEEEbiography}

\begin{IEEEbiography}[{\includegraphics[width=1in,height=1.25in,clip,keepaspectratio]{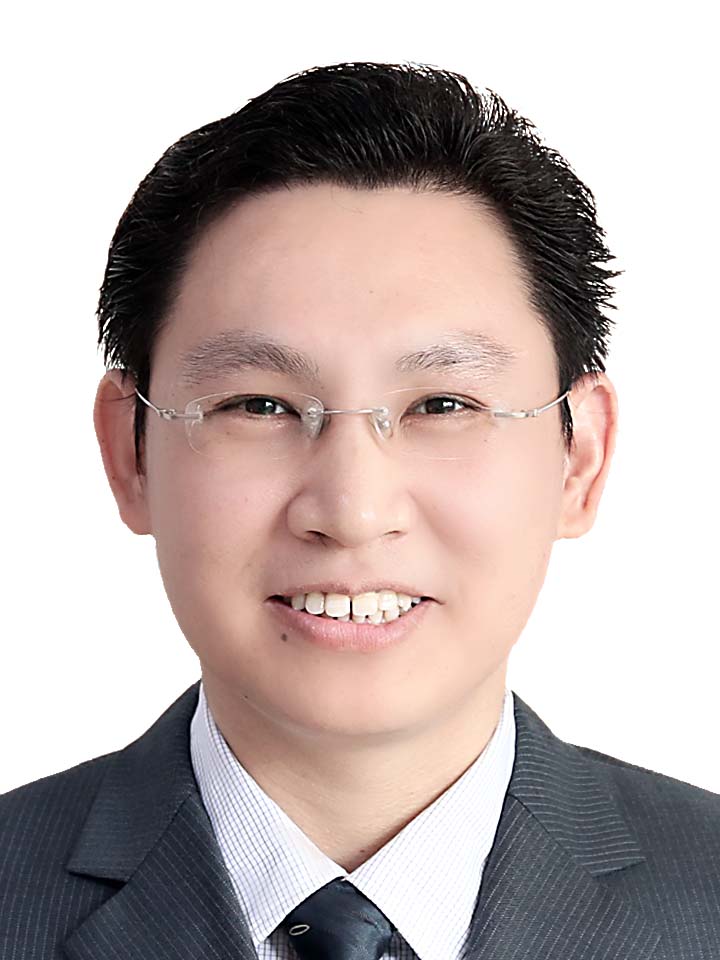}}]{Jian Yao}
 received the B.Sc. degree in Automation in 1997 from Xiamen University, P.R. China, the M.Sc. degree in Computer Science from Wuhan University, P.R. China, and the Ph.D. degree in Electronic Engineering in 2006 from The Chinese University of Hong Kong. From 2001 to 2002, he has ever worked as a Research Assistant at Shenzhen R$\&$D Centre of City University of Hong Kong. From 2006 to 2008, he worked as a Postdoctoral Fellow in Computer Vision Group of IDIAP Research Institute, Martigny, Switzerland. From 2009 to 2011, he worked as a Research Grantholder in the Institute for the Protection and Security of the Citizen, European Commission - Joint Research Centre (JRC), Ispra, Italy. From 2011 to 2012, he worked as a Professor in Shenzhen Institutes of Advanced Technology (SIAT), Chinese Academy of Sciences, P.R. China. Since April 2012, he has been a "Chutian Scholars Program" Distinguished Professor with School of Remote Sensing and Information Engineering, Wuhan University, P.R. China, and the director of Computer Vision and Remote Sensing (CVRS) Lab (CVRS Website: http://cvrs.whu.edu.cn/). He has published over 70 papers in international journals and proceedings of major conferences and is the inventor of more than 10 patents. His current research interests mainly include Computer Vision, Machine Vision, Image Processing, Pattern Recognition, Machine Learning, LiDAR Data Processing, SLAM, Robotics, etc.
\end{IEEEbiography}


\end{document}